\documentclass[sigconf]{acmart}

\usepackage{multirow}
\allowdisplaybreaks
\newtheorem{example}{Example}
\usepackage{enumerate}
\usepackage{balance}

\AtBeginDocument{%
  \providecommand\BibTeX{{%
    \normalfont B\kern-0.5em{\scshape i\kern-0.25em b}\kern-0.8em\TeX}}}

\copyrightyear{2023} 
\acmYear{2023} 
\setcopyright{acmlicensed}
\acmConference[SIGIR '23] {Proceedings of the 46th International ACM SIGIR Conference on Research and Development in Information Retrieval}{July 23--27, 2023}{Taipei, Taiwan.}
\acmBooktitle{Proceedings of the 46th International ACM SIGIR Conference on Research and Development in Information Retrieval (SIGIR '23), July 23--27, 2023, Taipei, Taiwan}
\acmPrice{15.00}
\acmISBN{978-1-4503-9408-6/23/07} 
\acmDOI{10.1145/3539618.3591667}

\begin{document}

\title{Distilling Semantic Concept Embeddings from Contrastively Fine-Tuned Language Models}


\author{Na Li}
\affiliation{%
  \institution{University of Shanghai for Science and Technology}
  \country{China} 
}
\email{li_na@usst.edu.cn}

\author{Hanane Kteich}
\affiliation{%
  \institution{CRIL CNRS \& University of Artois}
  \country{France} 
}
\email{kteich@cril.fr}

\author{Zied Bouraoui}
\affiliation{%
  \institution{CRIL CNRS \& University of Artois}
  \country{France} 
}
\email{bouraoui@cril.fr}

\author{Steven Schockaert}
\affiliation{%
  \institution{Cardiff University}
  \country{United Kingdom} 
}
\email{schockaerts1@cardiff.ac.uk}




\begin{abstract}
Learning vectors that capture the meaning of concepts remains a fundamental challenge. Somewhat surprisingly, perhaps, pre-trained language models have thus far only enabled modest improvements to the quality of such \emph{concept embeddings}. Current strategies for using language models typically represent a concept by averaging the contextualised representations of its mentions in some corpus. This is potentially sub-optimal for at least two reasons. First, contextualised word vectors have an unusual geometry, which hampers downstream tasks. Second, concept embeddings should capture the semantic properties of concepts, whereas contextualised word vectors are also affected by other factors. To address these issues, we propose two contrastive learning strategies, based on the view that whenever two sentences reveal similar properties, the corresponding contextualised vectors should also be similar. One strategy is fully unsupervised, estimating the properties which are expressed in a sentence from the neighbourhood structure of the contextualised word embeddings. The second strategy instead relies on a distant supervision signal from ConceptNet. Our experimental results show that the resulting vectors substantially outperform existing concept embeddings in predicting the semantic properties of concepts, with the ConceptNet-based strategy achieving the best results. These findings are furthermore confirmed in a clustering task and in the downstream task of ontology completion.
\end{abstract}

\ccsdesc[500]{Computing methodologies~Lexical semantics}
\ccsdesc[500]{Computing methodologies~Ontology engineering}
\ccsdesc[500]{Computing methodologies~Unsupervised learning}

\keywords{word embedding, language models, contrastive learning, commonsense knowledge}



\maketitle

\section{Introduction}
Since the introduction of BERT \cite{devlin-etal-2019-bert}, the focus in Natural Language Processing (NLP) has been on fine-tuning and exploiting large pre-trained language models, especially for solving sentence and paragraph level tasks. However, accurately modelling the meaning of individual words, in the form of static (i.e.\ not contextualised) vectors, also continues to be an important challenge. Static word vectors are used, among others, as pre-trained label embeddings for zero-shot \cite{DBLP:journals/corr/abs-1301-3666,ma-etal-2016-label} and few-shot learning \cite{DBLP:conf/nips/XingROP19,DBLP:conf/aaai/YanZH0BJS22,xiong-etal-2019-imposing,hou-etal-2020-shot,DBLP:conf/cvpr/Li0LFLW20,DBLP:conf/mir/YanBWJS21}; as concept representations for ontology alignment \cite{kolyvakis-etal-2018-deepalignment}, ontology completion \cite{DBLP:conf/semweb/LiBS19} and taxonomy learning \cite{DBLP:conf/sigir/VedulaNADS018,DBLP:conf/pkdd/MalandriMMN21}; for lexical substitution \cite{DBLP:conf/coling/0001B0L22} and topic modelling \cite{das-etal-2015-gaussian,DBLP:journals/tacl/DiengRB20,DBLP:journals/ipm/ZhaoWZLLZ21}; and for analysing social biases \cite{bommasani-etal-2020-interpreting}. Motivated by such applications, this paper focuses on representations of \emph{concepts}, rather than named entities. 

The distributional hypothesis \cite{harris1954distributional,firth1957synopsis} suggests that the meaning of a concept can be inferred from the contexts in which it appears. Standard word embedding models \cite{mikolov-etal-2013-linguistic,pennington-etal-2014-glove} implement this idea by using bag-of-words representations of these contexts. Clearly, such representations can only capture what is revealed about a concept in a very approximate way. Pre-trained language models (LMs), on the other hand, are able to capture meaning at the sentence level. LMs should thus enable us to obtain higher-quality context representations, which we would expect to translate into higher-quality concept embeddings. In particular, several authors have explored the idea that embeddings of concepts can be obtained by aggregating the contextualised embeddings of their mentions in some corpus \cite{ethayarajh-2019-contextual,bommasani-etal-2020-interpreting,vulic-etal-2020-probing,DBLP:conf/ijcai/LiBCAGS21,DBLP:journals/corr/abs-2012-15197,gupta-jaggi-2021-obtaining,DBLP:journals/taslp/WangCZ21}. While improvements over standard word embeddings are routinely reported, such improvements tend to be relatively small, and they are not always consistent. 

There are at least two challenges when it comes to learning concept embeddings in this way. First, contextualised word vectors are highly anisotropic \cite{ethayarajh-2019-contextual}. For unsupervised sentence embeddings, strategies aimed at reducing anisotropy have been found to result in substantial performance gains \cite{li-etal-2020-sentence,huang-etal-2021-whiteningbert-easy,liu-etal-2021-fast}. We may thus expect that concept embeddings can similarly benefit from such strategies. Second, and more fundamentally, contextualised word vectors do not only capture information about the meaning of words but also about their syntactic role and other characteristics of the sentences in which they appear \cite{DBLP:conf/iclr/TenneyXCWPMKDBD19,hewitt-manning-2019-structural,mickus-etal-2020-mean,luo-etal-2021-positional,timkey-van-schijndel-2021-bark}. If we are  interested in modelling the meaning of concepts, it thus seems beneficial to specialise the contextualised word vectors towards this aspect. Ideally, two contextualised word vectors should be similar if the corresponding sentences express similar properties, and dissimilar otherwise. This key idea is illustrated in the following example.
\begin{example}\label{ex1}
Consider the following sentences\footnote{All sentences were taken from GenericsKB \cite{DBLP:journals/corr/abs-2005-00660}.}:
\begin{enumerate}[(i)]
\itemsep0em
\item \emph{\underline{Submarines} can hide under the water.}
\item \emph{Some \underline{submarines} run on diesel engines.}
\item \emph{Some \underline{sharks} live at the bottom of deep underwater canyons.}
\item \emph{\underline{Trucks} are used to transport people or things, they use fuel known as diesel.}
\end{enumerate}
We would like the contextualised representation of \emph{submarines} in sentence (i) to be similar to the contextualised representation of \emph{sharks} in sentence (iii), as both sentences assert that the target concept has the property of being underwater. Similarly, we would like the representation of \emph{submarines} in sentence (ii) to be similar to the representation of \emph{trucks} in sentence (iv).
\end{example}
If we are able to learn contextualised word vectors that focus on the semantic properties that are expressed in a given sentence, we should be able to learn high-quality concept embeddings by averaging these contextualised representations across different sentences. 

In this paper, we propose and analyse a number of strategies based on contrastive learning to address the two aforementioned issues. Contrastive learning has already been successfully used for alleviating the anisotropy of BERT-based word and sentence embeddings \cite{gao-etal-2021-simcse,liu-etal-2021-fast}, based on the idea that embeddings of corrupted inputs should be similar to embeddings of the original word or sentence. Different from these approaches, our motivation for using contrastive learning is to move contextualised word vectors that capture similar semantic properties closer together, while vectors capturing different properties are pushed further apart. 

Crucially, to implement this idea, we need examples of sentences that express similar properties. We propose two strategies for identifying such sentences. Our first strategy is purely unsupervised. The main idea is to rely on the neighbourhood structure of standard contextualised word vectors. First note that when obtaining contextualised word vectors, we mask the target concept, following \cite{DBLP:conf/ijcai/LiBCAGS21}. This ensures that contextualised word vectors reflect the sentence context of the given concept, rather than any prior knowledge about the concept that is captured by the language model itself. Now suppose we have a contextualised representation of \textit{submarine}, and we look for the most similar contextualised word vectors, across a given corpus. Since the target concept is masked, these vectors may correspond to different words. Suppose, for instance, that they correspond to the words \textit{car}, \textit{truck} and \textit{airplane}. Then we can intuitively assume that the given sentence expresses the property of being a vehicle. Based on this idea, we can identify sentences that are likely to express the same property. Our second strategy uses a form of distant supervision, using knowledge about the commonsense properties of concepts from ConceptNet \cite{DBLP:conf/aaai/SpeerCH17}. For example, ConceptNet contains the triple \emph{(gun,HasProperty,dangerous)}. Given this triple, if a sentence contains both the words \emph{gun} and \emph{dangerous}, we assume it expresses that guns are dangerous. For each property encoded in ConceptNet, we can thus find sentences which express that the target concept has that property. This, in particular, allows us to find sentences that express the same property. 

We experimentally compare the concept embeddings that are obtained with the two aforementioned strategies. We are specifically interested in the extent to which different kinds of semantic properties can be predicted from these embeddings. We also evaluate our embeddings in a clustering task and an ontology completion task \cite{DBLP:conf/semweb/LiBS19,DBLP:journals/corr/abs-2202-09791}.  For both strategies, we find that our concept embeddings consistently outperform existing models by a substantial margin.

\section{Related Work}
The use of pre-trained language models for generating static word embeddings has already been extensively explored. A popular strategy is to aggregate the contextualised representation of a word $w$ across a number of sentences mentioning this word \cite{ethayarajh-2019-contextual,bommasani-etal-2020-interpreting,vulic-etal-2020-probing}. 
Several variations of this strategy have been studied, which mostly differ in how the contextualised representation of $w$ is computed. It is common to use the representation from the final layer of the transformer model or to average the representations from the final four layers, while \citet{vulic-etal-2020-probing} suggested averaging the first $k$ layers, with the optimal $k$ depending on the task. For words that consist of multiple tokens, the representations of these tokens are typically averaged. To aggregate the contextualised representations of a given word $w$ across multiple sentences, the most common strategy is to simply average them. \citet{ethayarajh-2019-contextual} instead proposed to take the first principal component, which produces almost the same result, given that the contextualised vectors are all located in a very narrow cone. 
In this paper, we build on the approach from \citet{DBLP:conf/ijcai/LiBCAGS21}, which masks the target word $w$ and uses the contextualised representation of the mask token; this approach is discussed in more detail in the next section. Beyond averaging-based strategies, some approaches have been inspired by Word2Vec \cite{mikolov-etal-2013-linguistic} or GloVe \cite{pennington-etal-2014-glove}, relying on BERT to obtain context embeddings \cite{gupta-jaggi-2021-obtaining,DBLP:journals/taslp/WangCZ21}, or to generate synthetic co-occurrence counts \cite{DBLP:journals/corr/abs-2012-15197}. 

Instead of relying on words in context, some approaches simply feed the word $w$ to the language model. \citet{bommasani-etal-2020-interpreting} found this to perform poorly with pre-trained models. However, better results were reported by \citet{vulic-etal-2021-lexfit}, after fine-tuning the BERT encoder on synonymy and antonymy pairs. 
\citet{gajbhiye-etal-2022-modelling} jointly fine-tuned a BERT encoder for concepts and an encoder for properties, using hypernyms from Microsoft Concept Graph \cite{DBLP:journals/dint/JiWSZWY19} and sentences from GenericsKB \cite{DBLP:journals/corr/abs-2005-00660} as training data.
MirrorBERT \cite{liu-etal-2021-fast} is a BERT encoder for both words and sentences, which is trained in a fully self-supervised way. It uses dropout to generate different variants of the same input, and then fine-tunes BERT such that these variants are closer to each other than to encodings of other inputs. The resulting encoder can generate high-quality word vectors, again without needing sentences mentioning the word in context. MirrorWiC \cite{liu-etal-2021-mirrorwic} can be seen as an adaptation of the MirrorBERT strategy to words in context. In particular, given a sentence $s$ mentioning some word $w$, multiple encodings of $w$ are obtained by (i) randomly masking different spans in $s$ and (ii) using dropout. The model then encourages different encodings of same sentence to be closer to each other than to encodings obtained from different sentences (even if the target word $w$ is the same). 

The aforementioned approaches have been developed with different tasks in mind. While word similarity benchmarks remain a popular choice for evaluating word vectors, \citet{DBLP:conf/ijcai/LiBCAGS21} and \citet{gajbhiye-etal-2022-modelling} were specifically interested in predicting the commonsense properties of concepts, while \citet{liu-etal-2021-mirrorwic} focused on word sense disambiguation. Accordingly, some of these approaches have complementary strengths. For instance, the model from \citet{DBLP:conf/ijcai/LiBCAGS21} outperformed the baselines on concept categorisation tasks, but under-performed in word similarity. In terms of downstream applications, since the introduction of BERT, word embeddings have primarily been used in settings where word meaning has to be modelled in the absence of any sentence context. For instance, word embeddings have been used to estimate class prototypes for few-shot learning, e.g.\ in image classification \cite{DBLP:conf/nips/XingROP19,DBLP:conf/aaai/YanZH0BJS22,DBLP:conf/mir/YanBWJS21} and for slot tagging in dialogue systems \cite{hou-etal-2020-shot}. In \cite{DBLP:conf/cvpr/Li0LFLW20}, word vector similarity was used to set an adaptive margin, as part of a margin-based model for few-shot image classification, to capture the idea that image classes with similar labels can be harder to differentiate.
Word embeddings have also been used for modelling label dependencies in multi-label classification \cite{xiong-etal-2019-imposing}.  Furthermore, word vectors have been used for ontology engineering tasks, e.g.\ for aligning ontologies \cite{kolyvakis-etal-2018-deepalignment} or for inferring plausible rules \cite{DBLP:conf/semweb/LiBS19}. In such applications, what matters is that concepts with similar word vectors have similar properties. We will focus on ontology completion in more detail in Section \ref{secOntologyCompletion}. In other applications, what matters is rather that clusters of word vectors are semantically coherent, e.g.\ when using word vectors for learning taxonomies \cite{DBLP:conf/sigir/VedulaNADS018,DBLP:conf/pkdd/MalandriMMN21} or for topic modelling \cite{das-etal-2015-gaussian,DBLP:journals/tacl/DiengRB20,DBLP:journals/ipm/ZhaoWZLLZ21}. Word vectors are much easier to train than language models, and can thus more easily be adapted. This advantage has been exploited to learn personal word embeddings, as part of a system for personalised search \cite{DBLP:journals/tois/YaoDW22}, or for studying how word meaning changes over time \cite{kutuzov-etal-2018-diachronic}. Finally, some authors have found that even for tasks where we need to model the meaning of words in context, using static word vectors can sometimes be beneficial \cite{alghanmi-etal-2020-combining,liu-etal-2020-towards,DBLP:conf/coling/0001B0L22}.

\section{Distilling Concept Embeddings}\label{secBackground}
In this section, we recall the concept embedding strategy from \citet{DBLP:conf/ijcai/LiBCAGS21}, which uses a pre-trained BERT model. The aim of our paper is to analyse how better concept embeddings can be obtained by instead relying on a suitably fine-tuned BERT model. Our proposed fine-tuning strategies will be the focus of Section \ref{secContrastive}.

Let $s_1,...,s_n$ be sentences in which some concept $c$ is mentioned. To obtain a vector representation of $c$ from the sentence $s_i$, \citet{DBLP:conf/ijcai/LiBCAGS21} replace $c$ by the \textit{<mask>} token and take the final-layer contextualised representation of this token, using a BERT-based language model. By masking the concept $c$, the resulting vector intuitively captures what the sentence $s_i$ reveals about the meaning of $c$, rather than any prior knowledge about the meaning of $c$ that is encoded in the language model itself. They found that this masking strategy improves how well the resulting embeddings capture the semantic properties of concepts. Let $\mathbf{x_1},...,\mathbf{x_n}$ be the vectors that are thus obtained from the available sentences. We refer to these vectors as the \emph{mention vectors} of concept $c$. We write $\mu(c)=\{\mathbf{x_1},...,\mathbf{x_n}\}$ for the set of mention vectors associated with $c$.
An embedding of concept $c$ can be obtained by averaging these mention vectors:
$$
\mathbf{c} = \frac{1}{|\mu(c)|} \sum\{ \mathbf{x} \mid \mathbf{x}\in \mu(c)\}
$$
However, not all sentences are equally informative. \citet{DBLP:conf/ijcai/LiBCAGS21} in particular highlighted issues that arise when sentences use concepts in idiosyncratic ways. For instance, sentences about the children's song ``Mary had a little lamb'' are unlikely to be useful for learning a representation of the concept \emph{lamb}. To reduce the impact of such idiosyncratic sentences, they proposed the following filtering strategy. Let $V$ be a vocabulary of concepts and let $M=\bigcup_{v\in V} \mu(v)$ be the set of all mention vectors, across all words in the vocabulary. For each mention vector $\mathbf{x}$ in $\mu(c)$, we compute its $k$ nearest neighbours among the vectors in $M$. If all $k$ of these neighbours belong to $\mu(c)$, $\mathbf{x}$ is deemed to be idiosyncratic. The embedding of concept $c$ is then obtained by averaging the remaining mention vectors, after removing the idiosyncratic ones.  The underlying intuition is based on the idea that the mention vectors in $\mu(c)$ capture the properties of $c$. If all the neighbours of such a mention vector $\mathbf{x}$ are associated with $c$, it suggests that the property which is captured by $\mathbf{x}$ only applies to that concept and is thus unlikely to be important. 

\section{Contrastive Learning Strategies}\label{secContrastive}
Each mention vector in $\mu(c)$ intuitively encodes what the corresponding sentence reveals about the concept $c$. It would thus be desirable if two mention vectors were similar if and only if the corresponding sentences reveal similar properties. Unfortunately, this is not always the case, given that contextualised vectors are affected by aspects such as word position, word frequency, and punctuation \cite{mickus-etal-2020-mean,luo-etal-2021-positional,timkey-van-schijndel-2021-bark}, which are irrelevant to word meaning, as well as the syntactic role of a word \cite{DBLP:conf/iclr/TenneyXCWPMKDBD19,hewitt-manning-2019-structural}, which is only loosely related. Our solution is to fine-tune the mention vectors using a contrastive learning strategy. While contrastive learning is a popular representation learning technique, it is usually applied in a purely unsupervised setting. For instance, to learn sentence embeddings using contrastive learning, one usually trains the model such that embeddings of corrupted versions of the same sentence are similar to each other, and dissimilar from embeddings of other sentences \cite{gao-etal-2021-simcse,liu-etal-2021-fast}. The same strategy has been used in \cite{liu-etal-2021-fast} for obtaining word embeddings from BERT. While it leads to embeddings that perform well on word similarity benchmarks, as we will see in our experiments, they are less suitable for tasks such as ontology completion, where we need concept embeddings that capture the semantic properties of the corresponding concepts.

In contrast to these existing approaches, our strategies will rely on weakly labelled training examples. Each example consists of two sentence-concept pairs, $(s_1,c_1)$ and $(s_2,c_2)$, where $c_i$ is a concept that is mentioned in sentence $s_i$. For positive training examples, the assumption is that the property that sentence $s_1$ expresses about concept $c_1$ is the same as what sentence $s_2$ expresses about $c_2$. For instance, if we write $s_{(i)}$ for sentence $(i)$ from Example \ref{ex1}, and similar for $s_{(ii)}$ and $s_{(iii)}$, then $(s_{i},\textit{submarines}); (s_{iii},\textit{sharks})$ could be a positive training example, while $(s_{i},\textit{submarines}); (s_{ii},\textit{submarines})$ could be a negative example. To implement our strategy, we thus first need to find a way to obtain such weakly labelled training examples. In Section \ref{secPositive} we propose two solutions for this problem: an unsupervised strategy which relies on the neighbourhood structure of the mention vectors, and a distantly supervised strategy which is based on ConceptNet. In Section \ref{secFinetuning} we then describe how the resulting training examples can be used for fine-tuning the model. 

\subsection{Constructing Weakly Labelled Examples}\label{secPositive}
We propose two strategies for obtaining weakly labelled training examples. These examples will then be used in Section \ref{secFinetuning} for fine-tuning the mention vectors.

\subsubsection{Neighbourhood Structure}\label{secNeighbourhood}
Consider sentences (i) and (iii) from Example \ref{ex1}. Even though these sentences express a similar property (i.e.\ being located under water), the resulting mention vectors are not actually similar, even after masking the target concepts. In fact, this is precisely our motivation for fine-tuning the mention vectors. To discover sentences which are likely to express a similar property, it is thus not sufficient to directly compare the corresponding mention vectors. Let us write $\phi(s,c)$ for the mention vector which is obtained after masking concept $c$ in sentence $s$. Essentially, two mention vectors $\phi(s_1,c_1)$ and $\phi(s_2,c_2)$ are similar if the following two conditions are satisfied for the sentences $s_1$ and $s_2$: (i) they express a similar property about their target concepts (i.e.\ $c_1$ and $c_2$) and (ii) they have a similar structure, with $c_1$ and $c_2$ moreover occurring in a similar syntactic role. In particular, if two mention vectors are similar, it is likely that they capture a similar property, even if the converse is not true. This insight can be used to compare the mention vectors $\phi(s_1,c_1)$ and $\phi(s_2,c_2)$ in an indirect way: we obtain the set $X_1$ of mentions vectors which are most similar to $\phi(s_1,c_1)$ and the set $X_2$ of mention vectors which are most similar to $\phi(s_2,c_2)$. If the concepts associated with the mention vectors in $X_1$ are broadly the same as the concepts associated with the mention vectors in $X_2$, it intuitively means that the property expressed by the vector $\phi(s_1,c_1)$ applies to the same set of concepts as the property expressed by the vector $\phi(s_2,c_2)$. In such a case, it is likely that $\phi(s_1,c_1)$ and $\phi(s_2,c_2)$ express the same property.

We now describe the proposed method more formally. Let $V$ be the vocabulary of all concepts and let $M=\bigcup_{c\in V} \mu(c)$ be the set of available mention vectors. In the following, we will assume that $\mu(c)\cap\mu(d)=\emptyset$ for $c\neq d$, i.e.\ we never have the exact same mention vector for different concepts. This assumption simplifies the formulations and is satisfied in practice. In particular, we can then link each mention vector $\mathbf{x}\in M$ to its unique corresponding concept, which we denote by  $\omega(\mathbf{x})$, i.e.\ we have $\omega(\mathbf{x}) = c$ iff $\mathbf{x}\in \mu(c)$. For a mention vector $\mathbf{x}\in M$, we write $\textit{neigh}(\mathbf{x})$ for its $k$ nearest neighbours from $M$, in terms of cosine similarity. Our central assumption is that when two mention vectors $\mathbf{x}$ and $\mathbf{y}$ express a similar property, then the concepts associated with the mention vectors in $\textit{neigh}(\mathbf{x})$ and $\textit{neigh}(\mathbf{y})$ will be similar. Formally, we define the compatibility degree $\pi(\mathbf{x},\mathbf{y})$ between $\mathbf{x}$ and $\mathbf{y}$ as follows:
\begin{align*}
\frac{\sum_{c\in V} \min(\textit{freq}(c,\textit{neigh}(\mathbf{x})),\textit{freq}(c,\textit{neigh}(\mathbf{y}))}{\sum_{c\in V}\max(\textit{freq}(c,\textit{neigh}(\mathbf{x})),\textit{freq}(c,\textit{neigh}(\mathbf{y}))}
\end{align*}
where $\textit{freq}(c,X) = |\{ \mathbf{x}\in X : \omega(\mathbf{x})=c\}|$ is the number of mention vectors in $X$ that are associated with concept $c$.
The following toy example provides an illustration of how $\pi(\mathbf{x},\mathbf{y})$ is computed.

\begin{example}
Figure \ref{figNeighbourBasedSelection} focuses on mention vectors $\mathbf{x}$, $\mathbf{y}$ and $\mathbf{z}$, along with their $k=4$ nearest neighbours. While $\mathbf{x}$ and $\mathbf{y}$ are not similar, their neighbours correspond to similar words. We have $\textit{neigh}(\mathbf{x})=\{\mathbf{x_1},\allowbreak\mathbf{x_2},\allowbreak\mathbf{x_3},\allowbreak\mathbf{x_4}\}$ and $\textit{neigh}(\mathbf{y})=\{\mathbf{y_1},\allowbreak\mathbf{y_2},\allowbreak\mathbf{y_3},\allowbreak\mathbf{y_4}\}$. We thus find:
\begin{align*}
\textit{freq}(\text{diver},\textit{neigh}(\mathbf{x})) &= 1 &
\textit{freq}(\text{diver},\textit{neigh}(\mathbf{y})) &= 1\\
\textit{freq}(\text{shark},\textit{neigh}(\mathbf{x})) &= 1&
\textit{freq}(\text{shark},\textit{neigh}(\mathbf{y})) &= 1\\
\textit{freq}(\text{submarine},\textit{neigh}(\mathbf{x})) &= 1&
\textit{freq}(\text{submarine},\textit{neigh}(\mathbf{y})) &= 0\\
\textit{freq}(\text{coral},\textit{neigh}(\mathbf{x})) &= 1&
\textit{freq}(\text{coral},\textit{neigh}(\mathbf{y})) &= 2
\end{align*}
with the frequencies for all other concepts being 0.
We thus obtain:
\begin{align*}
\pi(\mathbf{x},\mathbf{y}) = \frac{1+1+0+1}{1+1+1+2} = \frac{3}{5}
\end{align*}
As $\pi(\mathbf{x},\mathbf{y})$ is rather high, we will aim to move $\mathbf{x}$ and $\mathbf{y}$ closer together. In particular, $\mathbf{x}$ should be closer to $\mathbf{y}$ than to $\mathbf{z}$, despite the fact that  $\mathbf{x}$ and $\mathbf{z}$ correspond to the same word.

\begin{figure}
\includegraphics[width=200pt]{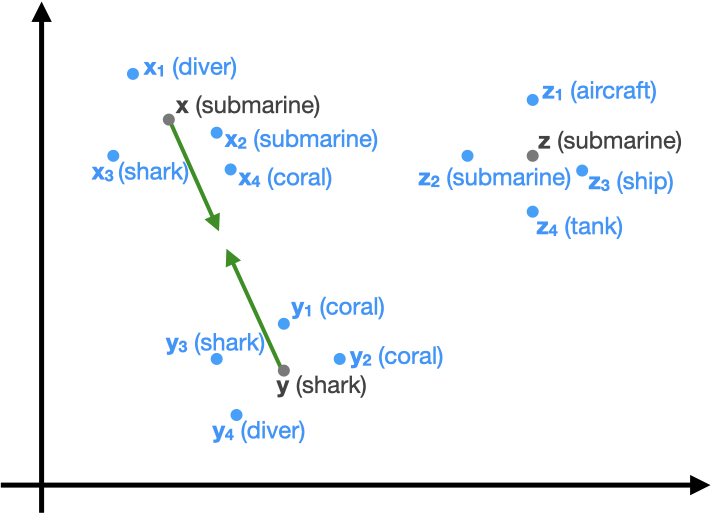}
\caption{Illustration of the neighbourhood-based selection of positive examples. \label{figNeighbourBasedSelection}}
\Description{The figure shows a two dimensional embedding of some mention vectors. There are two mention vectors which have a similar neighbourhood, namely $x$ and $y$. Arrows in the figure indicate that these vectors should be moved closer together.}
\end{figure}
\end{example}
In the following, we write $\textit{Pos} \subseteq (S\times V) \times (S\times V)$ to denote the resulting set of positive examples. Note that the elements of $\textit{Pos}$ are pairs of sentence-concept pairs. In particular, we have:
$$
\textit{Pos} = \{((s_1,c_1),(s_2,c_2)) \,|\,  \pi(\phi(s_1,c_1),\phi(s_2,c_2))\geq \theta, s_1\neq s_2\}
$$
for some threshold $\theta >0$.

\subsubsection{Distant Supervision from ConceptNet}\label{secConceptNet}
We now consider a strategy which uses ConceptNet \cite{DBLP:conf/aaai/SpeerCH17} as a distant supervision signal to identify positive training examples. ConceptNet contains a large number of triples of the form ([concept], \emph{HasProperty}, [property]). We first collected all the concept-property pairs that appear in such triples. We then removed those concept-property pairs for which the property only appears for at most two concepts. Let $T$ be the resulting set of concept-property pairs. For each pair $(c,p)\in T$, we identified all sentences in Wikipedia that mention both the concept $c$ and the property $p$. We rely on the simplifying assumption that such sentences express the knowledge that concept $c$ has property $p$, similar to the standard assumption underpinning distant supervision strategies for relation extraction \cite{mintz-etal-2009-distant}.  
Let $S_p$ be the resulting set of sentence-concept pairs for property $p$, i.e.\ $(s,c)\in S_p$ if sentence $s$ mentions both the concept $c$ and some property $p$ such that $(c,p)\in T$. The set of positive examples is then defined as follows:
$$
\textit{Pos} = \{((s_1,c_1),(s_2,c_2))\,|\, \exists p \,.\,(s_1,c_1)\in S_p, (s_2,c_2)\in S_p, s_1\neq s_2\}
$$
In other words, $(s_1,c_1)$ and $(s_2,c_2)$ are treated as a positive example if (i) the sentences $s_1$ and $s_2$ mention the same property $p$ and (ii) the corresponding target concepts $c_1$ and $c_2$ have $p$ in ConceptNet.

\subsection{Fine-tuning Strategies}\label{secFinetuning}
We now describe how the positive examples that were identified in Section \ref{secPositive} can be used for fine-tuning the mention vectors. The most straightforward strategy, which we discuss in Section \ref{secFineTunedLoss}, is based on fine-tuning the language model itself. The main drawback of this method is that it is computationally expensive. For this reason, in Section \ref{secProjection} we first discuss a simpler strategy, which simply learns a linear projection of the standard mention vectors.

\subsubsection{Projection Method}\label{secProjection}
Our aim is to learn a projection matrix $\mathbf{A}\in \mathbb{R}^{m\times n}$ 
such that vectors $\mathbf{A}\phi(s_1,c_1)$  and $\mathbf{A}\phi(s_2,c_2)$ are similar iff $((s_1,c_1),(s_2,c_2))\in \textit{Pos}$. 
Here $n$ is the dimension of mention vectors while $m$ is the dimension of the resulting vectors. 
We can think of $\mathbf{A}$ as selecting the subspace of the mention vector space that is focused on semantic properties. 
We use the supervised contrastive loss from \citet{DBLP:conf/nips/KhoslaTWSTIMLK20} to learn $\mathbf{A}$.
Let $B \subseteq S\times V$ be the set of sentence-concept pairs that are considered in a given mini-batch.
Let $X_{(s,c)} = \{(s',c') \,|\, ((s,c),(s',c'))\in \textit{Pos} \cap (B\times B)\}$ be the set of positive examples for $(s,c)$ in the mini-batch.
The loss is as follows:
\begin{align*}
\sum_{(s,c)\in B} \frac{-1}{|X_{(s,c)}|}\sum_{(s',c')\in X_{(s,c)}} \log \frac{e^{\cos(\mathbf{A}\phi(s,c),\mathbf{A}\phi(s',c'))/\tau}}{\sum_{(s'',c'')} e^{\cos(\mathbf{A}\phi(s,c),\mathbf{A}\phi(s'',c''))/\tau}}
\end{align*}
where the summation in the denominator ranges over $(s'',c'')\in B\setminus\{(s,c)\}$, and  the temperature $\tau>0$ is a hyperparameter. 

\subsubsection{Fine-Tuning BERT}\label{secFineTunedLoss}
We now consider a variant in which the contrastive loss is used to fine-tune a BERT encoder. This should allow us to learn more informative mention vectors, but at a higher computational cost.  Let us write $\psi(s,c)$ for the encoding of sentence-concept pair $(s,c)$ according to the fine-tuned BERT encoder (to distinguish it from $\phi$, which uses the pre-trained language model). Let $B$ and $X_{(s,c)}$ be defined as before.
We use the following loss:
\begin{align*}
\sum_{(s,c)\in B} \frac{-1}{|X_{(s,c)}|}\sum_{(s',c')\in X_{(s,c)}} \log \frac{e^{\cos(\psi(s,c),\psi(s',c'))/\tau}}{\sum_{(s'',c'')} e^{\cos(\psi(s,c),\psi(s'',c''))/\tau}}
\end{align*}
where the summation in the denominator ranges over $(s'',c'')\in B\setminus\{(s,c)\}$, as before, and $\tau>0$ is again a hyperparameter.




\begin{table*}[t]
\centering
\small
\caption{Results ($\%$) for BERT-large-uncased on the lexical classification tasks, in terms of  F1 ($\%$).\label{tabClassificationBERT}}
\begin{tabular}{
l
cc
cc
cc
cc
cc
cc
cc
}
\toprule  
& \multicolumn{2}{c}{\textbf{X-McRae}}
& \multicolumn{2}{c}{\textbf{CSLB}}
& \multicolumn{2}{c}{\textbf{Morrow}}
& \multicolumn{2}{c}{\textbf{WNSS}}
& \multicolumn{2}{c}{\textbf{BabelDom}}
& \multicolumn{2}{c}{\textbf{BM}}
& \multicolumn{2}{c}{\textbf{AP}}\\
\cmidrule(lr){2-3}
\cmidrule(lr){4-5}
\cmidrule(lr){6-7}
\cmidrule(lr){8-9}
\cmidrule(lr){10-11}
\cmidrule(lr){12-13}
\cmidrule(lr){14-15}
& SVM & CNN        
& SVM & CNN
& SVM & CNN
& SVM & CNN
& SVM & CNN
& SVM & CNN
& SVM & CNN\\
 \midrule
GloVe  &    
63.6 & -&
42.7 &- &
57.1&- &
48.6&- &
41.9& - &
59.4&-&
60.7 &-
\\  
Skip-Gram    &     
61.3 & - &
50.2 & -&
64.7 & -&
55.9 & -&
49.3 & -&
60.3&-&
61.7 &-
\\

Word2Sense &  
52.3&-&
50.3&-&
69.2&-&
43.9&-&
32&-&
63.8&- &
62.1 & -\\

SynGCN &  
56.5&-&
50.9&-&
71.4&-&
42.3&-&
34.2&-&
76.2 &-&
75.6&-\\

Numberbatch&  
63.5&-&
57.8&-&
71.1&-&
63.4&-&
41.5&-&
80.7&-&
82.3\\

MirrorBERT   &   
63.3&-&
51.6&-&
69.8&-&
59.1&-&
50.3&-&
79.2&-&
82.8 &-\\ 

\midrule
MirrorWiC & 
64.2&67.6&
52.7&60.1&
70.6&79.3&
59.1&63.4&
50.4&56.2&
80.1&81.6&
81.4 &82.6\\ 

No-Mask & 
55.9 & 57.3 &
45.6 & 46.8 &
67.5 & 68.2 &
50.9 & 51.8 &
40.3 & 42.4&
67.2&68.4&
62.5 &64.1\\

Mask & 
62.8 & 66.8 &
44.8 & 47.2 & 
57.8 & 59.3 & 
56.5 & 57.3 & 
49.3 & 51.1 & 
78.6& 80.1&
79.3 &81.9\\

Mask + filtering &
64.1 & 67.7 &
51.4 & 54.3 &
73.5 & 75.4 &
58.5 & 61.3 &
50.9 & 53.6&
79.6&82.6&
81.9&82.3\\

\midrule
ConProj  & 
66.6 &	69.3&
53.6 &	61.4&
75.5 &	81.1&
63.2 &	65.8&
54.7 & 58.4 & 
80.6&82.7&
82.9 &83.8\\  

ConFT &
67.4 &	69.8 &
55.7 &	63.6 &
76.9 &	82.4 &
65.7 & 67.2 &
55.8  & 59.6 & 
81.1&82.9&
83.3 &84.2\\  

ConCN&  
68.3&70.9&
56.2&65.1&
77.5&83.8&
67.1&69.4&
57.3&61.7&
81.8&83.6&
84.1 &85.3\\

ConProj + filt.& 
70.1 &	73.2&
56.3 &	68.8&
78.8 &	83.7&
65.2 &	68.6&
59.1 & 63.9&
81.2&83.3&
83.4 &84.6\\

ConFT + filt.\ 
&
71.9 &	74.4 &
57.3 &	69.3 &
78.5 &	86.2 &
67.1 & 69.3 &
60.7  & 64.8 &
82.1&83.8&
84.1 &85.1\\  

ConCN + filt.\ &
\textbf{73.7} &	\textbf{75.2} &
\textbf{59.4} &	\textbf{71.8} &
\textbf{81.1} &	\textbf{87.5} &
\textbf{68.9} & \textbf{70.8} &
\textbf{62.5}  & \textbf{67.1} &
\textbf{83.2}&\textbf{84.7}&
\textbf{84.7} &\textbf{85.9}\\  

\bottomrule  
\end{tabular}
\end{table*}
\section{Experiments}\label{secExperiments}
We present an evaluation of our proposed strategies\footnote{Datasets and code at \url{https://github.com/lina-luck/semantic_concept_embeddings.}}. We will in particular focus on the following variants:
\begin{description}
\item[ConProj] uses the projection method for fine-tuning and the neighbourhood structure for obtaining positive examples.
\item[ConFT] fine-tunes the BERT encoder and uses the neighbourhood structure for obtaining positive examples.
\item[ConCN] fine-tunes the BERT encoder and uses the distant supervision strategy based on Conceptnet for obtaining positive examples.
\end{description}
By comparing these variants we are particularly interested in answering the following two research questions: (i) is learning a linear projection sufficient or do we need to fine-tune the language model, and (ii) how effective are the two proposed strategies for obtaining weakly labelled positive examples.
The primary focus of our experiments is on word classification (Section \ref{secWordClassification}), as these allow us to directly evaluate the extent to which our embeddings capture different kinds of semantic properties. This is motivated by the observation that this is precisely what matters in most applications where static concept embeddings are still needed. For instance, tasks such as ontology completion or zero-shot learning directly use concept embeddings to link concepts to their semantic properties. We also evaluate the quality of the clusters that arise from our embeddings (Section \ref{secClustering}). To verify that the concept embeddings are indeed useful in downstream applications, we present an evaluation on the downstream task of ontology completion (Section \ref{secOntologyCompletion}). We conclude  with an analysis of the main results (Section \ref{secAnalysis}).

\paragraph{Baselines}
We compare our embeddings with Skip-gram \cite{mikolov-etal-2013-linguistic} and GloVe \cite{pennington-etal-2014-glove}, as representative examples of traditional word embeddings\footnote{We used the Skip-gram embeddings trained on Google News (\url{https://code.google.com/archive/p/word2vec/}) and Glove embeddings trained on Common Crawl (\url{https://nlp.stanford.edu/projects/glove/}).}, and with SynGCN\footnote{\url{https://drive.google.com/file/d/1wYgdyjIBC6nIC-bX29kByA0GwnUSR9Hh/view}} \cite{vashishth-etal-2019-incorporating} and Word2Sense\footnote{\url{https://drive.google.com/file/d/1kqxQm129RVfanlnEsJnyYjygsFhA3wH3/view}} \cite{panigrahi-etal-2019-word2sense}, as examples of more recent static word embeddings. We furthermore compare with the Numberbatch\footnote{\url{https://conceptnet.s3.amazonaws.com/downloads/2019/numberbatch/numberbatch-en-19.08.txt.gz}} embeddings from \citet{DBLP:conf/aaai/SpeerCH17}, as these were also fine-tuned based on ConceptNet. Beyond traditional word embeddings, we compare with the method from \citet{DBLP:conf/ijcai/LiBCAGS21}, as we use their mention vectors as our starting point. We include two variants: one version where all mention vectors are averaged (\emph{Mask}) and one version where their filtering strategy is applied first (\emph{Mask+filtering}). In addition, we consider a variant in which mention vectors are obtained without masking the target concept (\emph{No-Mask}). In this case, for words that consist of more than one token, the contextualised token representations are averaged. Rather than taking the final layer representation, which has been found to be sub-optimal \cite{vulic-etal-2020-probing}, in this case, we select the optimal layer based on a validation split. Finally, we include results for MirrorBERT\footnote{\url{https://huggingface.co/cambridgeltl/mirror-bert-base-uncased-word}} \cite{liu-etal-2021-fast} and MirrorWiC\footnote{\url{https://huggingface.co/cambridgeltl/mirrorwic-bert-base-uncased}} \cite{liu-etal-2021-mirrorwic}, both of which also use a contrastively fine-tuned BERT model.

\paragraph{Training Details}
To obtain mention vectors, for each concept, we randomly sample up to 500 sentences mentioning that concept from Wikipedia. We use the same sentences for our methods, for the baseline methods from \citet{DBLP:conf/ijcai/LiBCAGS21} and for MirrorWiC.  Unless specified otherwise, we use BERT-large-uncased as the pre-trained language model. 
The learning rate for our models was set to 2e-4, with cosine warm-up for the first 2 epochs. We use early stopping with a patience of 10 and a minimum difference of 1e-10. We used the AdamW optimizer. We set the temperature parameter in the contrastive loss to 0.05 and the number of neighbours $k$ for evaluating the compatibility degree to 5. The threshold $\theta$ on compatibility degrees to be considered a positive example was set to 0.5. For implementing the contrastive loss, we relied on the Pytorch Metric Learning library\footnote{\url{https://kevinmusgrave.github.io/pytorch-metric-learning/}}. Based on the values reported by \citet{DBLP:conf/ijcai/LiBCAGS21}, we set the number of neighbours for the filtering strategy to 50 for X-McRae, WNSS and BabelDomains, and 5 for CSLB, Morrow,
BM and AP. The dimension $m$ of the transformed vectors, for the projection-based fine-tuning method, is 256. For \emph{ConProj}, we obtain the sentence-concept pairs for a given mini-batch by sampling 1024 such pairs from the set $\textit{Pos}$. For \emph{ConFT}, we proceed similarly, but limit the number of pairs to 512 due to memory constraints.
For \emph{ConCN}, the set of sentence-concept pairs for a given mini-batch is obtained by repeatedly (i) sampling a property $p$ and (ii) sampling 50 sentences from $S(p)$.

\subsection{Word Classification}\label{secWordClassification}
We consider a number of benchmarks which involve predicting whether a given concept belongs to some class, where the classes of interest correspond to different kinds of semantic properties, namely commonsense properties (e.g.\ being made of wood), taxonomic categories (e.g.\ being an animal) and thematic domains (e.g. related to music). We evaluate the extent to which these classes can be predicted from different kinds of concept embeddings. We have included the five benchmarks that were used by \citet{DBLP:conf/ijcai/LiBCAGS21}:
\begin{itemize}
\item the extension of the McRae feature norms \cite{mcrae2005semantic} that was introduced by \citet{DBLP:conf/cogsci/ForbesHC19} (X-McRae\footnote{\url{https://github.com/mbforbes/physical-commonsense}}), covering 513 words and 50 classes (being commonsense properties);
\item CSLB Concept Property Norms\footnote{\url{https://cslb.psychol.cam.ac.uk/propnorms}}, with 635 words and 395 classes (being commonsense properties);
\item the Morrow dataset \cite{morrow2005representation}, covering 888 words and 13 classes (being broad taxonomic categories such as \emph{animals});
\item WordNet supersenses\footnote{\url{https://wordnet.princeton.edu/download}} (WNSS), with 18200 words and 25 classes (being broad taxonomic categories);
\item BabelDomains\footnote{\url{http://lcl.uniroma1.it/babeldomains/}} \cite{camacho-collados-navigli-2017-babeldomains}, covering 12477 words and 28 classes (being thematic domains). 
\end{itemize}
For these datasets, we use the same training-tuning-test splits as \citet{DBLP:conf/ijcai/LiBCAGS21}\footnote{It should be noted that the annotations in CSLB are not complete, i.e.\ some properties which are not asserted to hold for a given concept are nonetheless valid \cite{misra2022property}. This means that care is needed when drawing conclusions from the absolute performance of models on this dataset. As we are mostly interested in the relative performance of different embeddings in this paper, this should not affect the analysis.}.
We also include two additional benchmarks\footnote{We used the versions available at \url{https://github.com/vecto-ai/word-benchmarks}.}:
\begin{itemize}
\item the Battig and Montague norms \cite{battig1969category}, with 5321 words and 56 classes (being fine-grained taxonomic categories such as \emph{weapon} or \emph{unit of time});
\item the dataset from \citet{almuhareb2005concept}, with 402 words and 21 classes (being WordNet hypernyms).
\end{itemize}
For both datasets, we randomly split the positive examples, for each category, into 60\% for training, 20\% for tuning and 20\% for testing. As these datasets only specify positive examples, for each concept, we generate 5 negative examples by randomly selecting categories to which the concept does not belong to.

\paragraph{Methodology}
For each class, we train a linear SVM to classify concepts based on their embedding. We report the results in terms of F1 score, macro-averaged across all classes from a given benchmark. We furthermore experiment with a simple Convolutional Neural Network (CNN), which takes the individual mention vectors as input, rather than their average. In particular, each mention vector is first fed through a dense layer and the resulting vectors are aggregated using max-pooling. This aggregated vector is then fed to a classification layer. For the SVM, we used the standard scikit-learn implementation. The C parameter is tuned from $\{0.1, 1, 10, 100\}$. For the CNN model, we have used the standard PyTorch implementation, setting the kernel size and stride to 1. We used 64 filters with ReLU activation, a batch size of 32 and a learning rate of 1e-3. The CNN is trained with binary cross-entropy, using Adam.

\paragraph{Results}
The results are summarised in Table \ref{tabClassificationBERT}. A number of clear observations can be made. First, all three of the proposed methods (\emph{ConProj}, \emph{ConFT}, \emph{ConCN}) outperform the baselines\footnote{Note that \emph{MirrorBERT} and \emph{MirrorWiC} use BERT-base, whereas our models and those from \citet{DBLP:conf/ijcai/LiBCAGS21} rely on BERT-large. However, as we will see below, the outperformance of our model remains after changing the encoder to BERT-base. We use BERT-large for the main experiments, as the methods from \citet{DBLP:conf/ijcai/LiBCAGS21}, which are our primary baselines, achieve substantially weaker results for BERT-base.}. The main exception is CSLB, where Numerbatch outperforms all SVM-based models apart from \emph{ConCN} with filtering. Among our proposed methods, \emph{ConCN} performs best, showing the effectiveness of the ConceptNet-based distant supervision strategy, while \emph{ConFT} outperforms \emph{ConProj}, as expected. As a second observation, the filtering strategy from \citet{DBLP:conf/ijcai/LiBCAGS21} is highly effective, offering improvements that are complementary to those of our proposed methods. Third, the CNN consistently outperforms the SVM model, with the margin being particularly large for CSLB.

\subsection{Clustering}\label{secClustering}
The BM \cite{battig1969category} and AP \cite{almuhareb2005concept} datasets, which we used for word classification, have also been used as clustering benchmarks in previous work \cite{baroni-etal-2014-dont}. Specifically, the aim is to organise the words from the dataset into semantically meaningful clusters. The clusters are evaluated using cluster purity, using the categories which are provided in the dataset as the ground truth. The main aim of this experiment is to analyse the quality of our embeddings in an unsupervised setting, to test their suitability for tasks such as topic modelling \cite{das-etal-2015-gaussian,DBLP:journals/tacl/DiengRB20,DBLP:journals/ipm/ZhaoWZLLZ21}. We use $k$-means to obtain the clusters, choosing $k$ as the number of categories from the dataset. Since the quality of the clusters is sensitive to the random initialisation of the clusters, we repeat the experiment 10 times and report the average purity.

The results are shown in Table \ref{tabClusteringOntologyCompletion}. As can be seen, our method outperforms all baselines. Similar as for word classification, we can see that ConCN is the best variant and that the filtering strategy consistently improves the results. Among the baselines, the strong performance of Numberbatch is also notable.  

\begin{table}[t]
\centering
\setlength{\tabcolsep}{2pt}
\small
\caption{Results for clustering and ontology completion using BERT-large-uncased. Clustering results are in terms of purity ($\%$) while ontology completion results are in terms of F1 ($\%$).\label{tabClusteringOntologyCompletion}}
\begin{tabular}{
l cc ccccc}
\toprule  
& \multicolumn{2}{c}{\textbf{Clustering}} & \multicolumn{5}{c}{\textbf{Ontology Completion}}\\
\cmidrule(lr){2-3}\cmidrule(lr){4-8}
& \textbf{BM}
& \textbf{AP}
& \textbf{Wine}
& \textbf{Econ}
& \textbf{Olym}
& \textbf{Tran}
& \textbf{SUMO}
\\
\midrule
Glove & 57.3	& 44.9   & 14.2 & 14.1 & 9.9 & 8.3 & 34.9 \\
Skip-Gram & 46.7 &	32.4 & 13.8 &  13.5 & 8.3 & 7.2 & 33.4\\ 
Word2Sense & 25.5	& 16.6 & 13.4 & 13.2 & 8.1 & 7.2 &  33.1 \\
SynGCN & 56.9 & 39.2 & 13.9 & 13.8 & 9.4 & 8.1 & 33.9\\
Numberbatch & 73.8 &	53.3 & 25.6 &26.2  &26.8& 16.0 &47.3\\
MirrorBERT & 62.4 &	51.4 & 22.5 & 23.8 & 20.9& 12.7 & 40.1 \\
MirrorWiC &64.6 & 52.5 &24.7& 24.9 & 22.1 & 13.9& 46.9 \\
Mask + filt.\ & 61.3 & 48.2 & 24.5 & 24.3 & 22.9& 13.0 & 46.4\\
\midrule
ConProj & 75.8 &  54.2 & 26.9 & 27.3 &25.6& 15.9 & 48.2\\
ConFT  & 76.1 & 56.9  & 27.5 & 29.2 & 26.5 & 17.4 & 48.6\\
ConCN & 76.9 & 57.2 &29.1 & 31.3 & 27.6 & 19.7& 50.4\\
ConProj + filt.\ & 76.3 & 54.9 & 27.2 & 28.6 &26.2& 17.1 & 49.3\\
ConFT + filt.\ & 76.8 & 57.3 & 28.7 & 30.3 & 28.2 & 19.1 & 50.3\\
ConCN + filt.\ & \textbf{77.4} & \textbf{57.9} &\textbf{31.3} & \textbf{32.4} & \textbf{29.7} & \textbf{20.9}& \textbf{52.6}\\
\bottomrule 
\end{tabular}
\end{table}
\subsection{Ontology Completion}\label{secOntologyCompletion}
An ontology can be viewed as a set of rules. A simple rule takes the following form:
$$
A_1(x)\wedge ... A_n(x) \rightarrow B(x)
$$
It expresses the knowledge that whenever some entity $x$ belongs to the concepts $A_1,...,A_n$ then it also belongs to the concept $B$. In general, rules may also contain constructs of the form $\exists y\,\, R(x,y) \wedge A(y)$, which expresses that $x$ is related, via relation $R$, to some instance of $A$. The key principle underpinning the ontology completion benchmarks from \cite{DBLP:conf/semweb/LiBS19} is that real-world ontologies often contain sets of closely related rules, which only differ in a single concept. Consider, for instance, an ontology containing the following rules:\\[-2em]

\begin{align*}
\textit{AppleJuice}(x) \wedge \textit{Small}(x) &\rightarrow \textit{SuitableForKids}(x)\\
\textit{PineappleJuice}(x) \wedge \textit{Small}(x) &\rightarrow \textit{SuitableForKids}(x)\\
\textit{MangoJuice}(x) \wedge \textit{Small}(x) &\rightarrow \textit{SuitableForKids}(x)
\end{align*}

\noindent For instance, the first rule intuitively captures the knowledge that a small portion of apple juice is suitable for kids to drink. From these rules, we may infer that the following rule should also be considered valid within the context of this ontology, even if it is not actually provided:\\[-2em]

\begin{align*}
\textit{OrangeJuice}(x) \wedge \textit{Small}(x) &\rightarrow \textit{SuitableForKids}(x)
\end{align*}

\noindent The underlying principle is that \emph{orange juice} satisfies all the properties that are common to \emph{apple juice}, \emph{pineapple juice} and \emph{mango juice}. To infer such plausible rules, we often need to combine prior knowledge about the meaning of the concepts (e.g.\ that orange juice has similar properties to apple juice and pineapple juice) with the knowledge that is inferred from the structure of the ontology itself (e.g.\ to deal with concepts whose name is not descriptive). To this end, \cite{DBLP:conf/semweb/LiBS19} introduced a graph neural network, in which the nodes correspond to concepts. Concepts that co-occur in the same rule are connected with an edge. The input representation of each node is a pre-trained concept embedding, which was taken to be the skip-gram embedding of the concept name in \cite{DBLP:conf/semweb/LiBS19}. Ontology completion has a number of practical applications. For instance, apart from suggesting plausible missing knowledge to ontology engineers, the ability to predict plausible rules also plays an important role in ontology alignment \cite{DBLP:conf/aaai/0008CA022}. 


Following, \cite{DBLP:conf/ijcai/LiBCAGS21}, we use ontology completion benchmarks for evaluating the quality of different types of concept embeddings, using the same methodology. In particular, we first tokenise concept names, based on the common naming conventions in ontologies. For instance, the concept \emph{PastaWithWhiteSauce} becomes ``pasta with white sauce''. If the resulting concept name does not appear in Wikipedia, we never predict this concept as a positive example. 
We use the same hyperparameters and training-test splits as \cite{DBLP:conf/semweb/LiBS19}, and use their evaluation scripts\footnote{\url{https://github.com/lina-luck/rosv_ijcai21}}. The benchmark includes five different ontologies. First, the SUMO ontology was included as a prototypical example of a large open-domain ontology. The other four are well-known domain-specific ontologies: Wine, Economy, Olympics and Transport\footnote{We used the training and test splits from \url{https://github.com/bzdt/GCN-based-Ontology-Completion}.}. 

The results for ontology completion in Table \ref{tabClusteringOntologyCompletion} are broadly in line with those from the word classification and clustering experiments. Note in particular how the performance of \emph{ConCN + filt}, our best-performing variant, is substantially higher than that of \emph{Numberbatch}, \emph{MirrorBERT}, \emph{MirrorWiC} and \emph{Mask + filt}, which in turn substantially outperform the remaining baselines. Overall, these results clearly show that high-quality concept embeddings can be extracted from language models, which have significant benefits over traditional word embeddings. For instance, with the exception of SUMO, all our methods achieve F1 scores which at least double the F1 scores of skip-gram. Moreover, compared to earlier BERT-based methods such as \emph{MirrorBERT}, \emph{MirrorWiC} and \emph{Mask}, our vectors are more focused on the semantic properties of concepts, which gives them a clear advantage in this task.

\begin{table}[t]
\centering
\small
\setlength{\tabcolsep}{1.7pt}
\caption{Comparison of different language models and strategies for selecting positive examples, for X-McRae, in terms of F1 ($\%$). Results are for BERT-base-uncased (BB), BERT-large-uncased (BL), RoBERTa-base (RB) and RoBERTa-large (RL).\label{tabAnalysisMC}}
\begin{tabular}{
l
cc
cc
cc
cc
}
\toprule  
& \multicolumn{2}{c}{\textbf{BB}}
& \multicolumn{2}{c}{\textbf{BL}}
& \multicolumn{2}{c}{\textbf{RB}}
& \multicolumn{2}{c}{\textbf{RL}}\\
\cmidrule(lr){2-3}
\cmidrule(lr){4-5}
\cmidrule(lr){6-7}
\cmidrule(lr){8-9}
& SVM & CNN        
& SVM & CNN
& SVM & CNN
& SVM & CNN\\

\midrule
No-Mask  &
49.6 & 51.2 &
55.9 & 57.3 &
50.4 & 52.4 &
53.5& 57.2 \\

Mask & 
53.6&  57.2 &
62.8 & 66.8  & 
52.1 & 54.6 & 
63.9 &  67.1 \\

Mask + filtering & 
58.2 & 60.3 &
64.1 & 67.7&
59.5 & 61.8&
64.8 &68.1 \\

\midrule

ConProj  & 
64.3& 67.6 &
66.6 & 69.3 &
64.9& 67.9& 
67.2 & 69.8  \\

ConFT  &   
65.2& 68.1 &
67.4 & 69.8 &
65.3& 68.2& 
67.4 & 70.1  \\

ConCN  & 
66.4& 69.5 &
68.3 &70.9 &
67.2& 70.0&
69.6 &71.3   \\

ConProj + filt.\  & 
66.3& 68.3 &
70.1 & 73.2 & 
67.7& 69.4 &
70.5 & 73.9    \\

ConFT + filt.\  & 
67.0& 70.1 &
71.9 & 74.4 &
68.2& 71.6 &
71.3 & 73.5 \\

ConCN + filt.\ & 
\textbf{68.3} & \textbf{72.5} &
\textbf{73.7} & \textbf{75.2} &
\textbf{69.1} & \textbf{73.3} &
\textbf{73.9} & \textbf{75.8}  \\
\midrule
W-ConProj  & 
61.2 & 65.9 &
64.9 & 68.7&
62.1 & 66.2&
65.9 & 69.2  \\ 

W-ConProj + filt.\  & 
63.8 & 67.3  &
68.6 & 71.9 &
64.5 & 69.2  &
70.1 & 73.6 \\

\bottomrule  
\end{tabular}
\end{table}

\begin{table*}[t]
\footnotesize
\centering
\caption{The table shows pairs of sentences whose mention vectors are similar when using the model fine-tuned with the ConCN strategy while being dissimilar when using pre-trained BERT model. \label{tabSimilarForFinetuned}}
\begin{tabular}{p{240pt}p{240pt}}
\toprule
\multicolumn{2}{c}{\textbf{Similar after fine-tuning but not for pre-trained model}}\\
\midrule
The second floor of the facade was originally designed to be a private Disney family \underline{apartment}. & It would also allow the RTC to buy new curtains and wall coverings, and to restore the \underline{building}'s exterior. \\[0.8em]

A vinaigrette can be made with black \underline{garlic}, sherry vinegar, soy, a neutral oil, and Dijon mustard.  & ... in flood situations where normal foods are out of reach black \underline{swans} will feed on pasture plants on shore.\\[0.8em]

... raising the value of the Icelandic \underline{crown} in 1925, very much as Winston Churchill raised the value of the pound ... & There are further plans to reintroduce the South African \underline{cheetahs} to the Lower Zambezi.\\
\bottomrule
\end{tabular}
\end{table*}


\begin{table}[t]
\caption{Nearest neighbours, in terms of cosine similarity, for some selected words, using WordNet supersenses vocabulary. \label{tabNNs}}
\footnotesize
\centering
\begin{tabular}{lll}
\toprule
 & \textbf{Word} & \textbf{Neighbours}\\
\midrule
\parbox[t]{2mm}{\multirow{10}{*}{\rotatebox[origin=c]{90}{\textbf{Numberbatch}}}} & lemon & citron, citrange, limeade, lime, lemonade \\
& deepening & broadening, deep, strengthening, deepness, worsening \\
& icon & iconology, symbol, iconography, iconoclasm, emblem \\
& stunt & trick, aerialist, jugglery, gimmickry, cartwheel\\
& milkman & dairyman, milk, creamery, clabber, lacteal\\
& paradox & antinomy, contradiction, duality, oxymoron, inconsistency \\
& desk & office, copyholder, desktop, bookcase, table \\
& beer & ale, brewery, microbrewery, brewpub, keg \\
& steam & steamer, steamboat, steamfitter, gasification, boiling \\
& razor & razorblade, shaver, blade, scissors, sharpener \\
\midrule
\parbox[t]{0mm}{\multirow{10}{*}{\rotatebox[origin=c]{90}{\textbf{MirrorBERT}}}} & lemon & lemonwood, lemonade, orangeade, limeade, dewberry \\
& deepening & deepness, broadening, deep, shallowness, diversification \\
& icon & iconoclast, iconography, iconoclasm, iconology, symbol \\
& stunt & trick, props, joyride, sabotage, leap\\
& milkman & dairyman, milk, alewife, grocer, milkwort \\
& paradox & contradiction, ambiguity, perplexity, singularity, unreality \\
& desk & office, clerk, bookcase, counter, receptionist \\
& beer & ale, liquor, rum, brewpub, brandy \\
& steam & steamfitter, boilerplate, turbine, generator, gasification \\
& razor & razorblade, blade, scissors, needle, knife \\
\midrule
\parbox[t]{0mm}{\multirow{10}{*}{\rotatebox[origin=c]{90}{\textbf{ConCN + filt.}}}} & lemon & lime, blueberry, tangerine, cranberry, lemonade \\
& deepening & broadening, weakening, mellowing, narrowing, depths \\
& icon & button, plaque, emblem, display, iconography \\
& stunt & handstand, gimmickry, skydiver, fling, skydiving \\
& milkman & cheesemonger, dairyman, barmaid, paperboy, cow \\
& paradox & singularity, irony, doublethink, unreality, perplexity \\
& desk & counter, sideboard, office, bookcase, drawer \\
& beer & mead, ale, vodka, brandy, tequila \\
& steam & electricity, furnace, turbine, vent, gasification \\
& razor & penknife, tool, scalpel, razorblade, shaver \\
\bottomrule
\end{tabular}
\end{table}

\subsection{Analysis}\label{secAnalysis}
We now present some additional analysis of our models, focusing primarily on the results for word classification.

\paragraph{Outperformance of the CNN} The CNN is expected to outperform when the semantic properties we need to predict are only rarely mentioned in text. Indeed, such properties will only be captured by a small number of mention vectors, and this information will be largely lost after averaging them. CSLB focuses on commonsense properties, many of which are indeed rarely expressed in text \cite{DBLP:conf/cikm/GordonD13}, which explains the large outperformance of the CNN model for this benchmark (as well as the comparatively strong performance of Numberbatch) in Table \ref{tabClassificationBERT}. For instance, the categories for which the difference in F1 score between the SVM and CNN models is largest, for \emph{ConCN+filtering}, are as follows: \emph{grows on plants}, \emph{is cool}, \emph{has a top}, \emph{is furry}, \emph{has green leaves}, \emph{is for soup}, \emph{is ridden}, \emph{is a body part}, \emph{is found in America}, \emph{has big eyes}, \emph{has arms}, \emph{has a blade/blades}. For X-McRae, the overall differences are smaller, which can be explained by the fact that several taxonomic properties are included in this dataset as well. However, for many commonsense properties, we similarly observe large differences in F1 score. The largest differences were observed for the following X-McRae properties: 
\emph{loud}, \emph{used for holding things}, \emph{words on it}, \emph{eaten in summer}, \emph{worn for warmth}, \emph{flies}, \emph{used for killing}, \emph{used for cleaning}, \emph{worn on feet}.

\paragraph{Comparing Language Models}
Table \ref{tabAnalysisMC} analyses the impact of changing the language model encoder, showing word classification results for BERT-base-uncased, BERT-large-uncased, RoBERTa-base and RoBERTa-large \cite{DBLP:journals/corr/abs-1907-11692}, for the SVM model. We can see that BERT-large and RoBERTa-large outperform the base models, as expected, but the differences for our methods are relatively small. In contrast, for the \emph{No-Mask}, \emph{Mask} and \emph{Mask+filtering} baselines, switching to the base models is more detrimental. Across all language models, we find that our proposed methods outperform the baselines. 

\paragraph{Importance of the Compatibility Degree}
For the \emph{ConProj} and \emph{ConFT} variants, the set of positive examples is based on the neighbourhood structure of the mention vectors (see Section \ref{secNeighbourhood}). Another possibility could be to simply assume that sentences mentioning the same word are likely to express the same property. In other words, we could define the set of positive examples as follows:
$$
\textit{Pos} = \{((s_1,c),(s_2,c)) \,|\,  s_1,s_2\in S, c\in V, s_1\neq s_2\}
$$
The effectiveness of this alternative strategy is analysed in Table \ref{tabAnalysisMC}, where it is referred to as \emph{W-ConProj} (when used in combination with the projection-based contrastive loss). While this alternative strategy also outperforms the baselines, it consistently underperforms our main neighbourhood-based strategy.

\begin{figure}
\centering
\includegraphics[width=200pt]{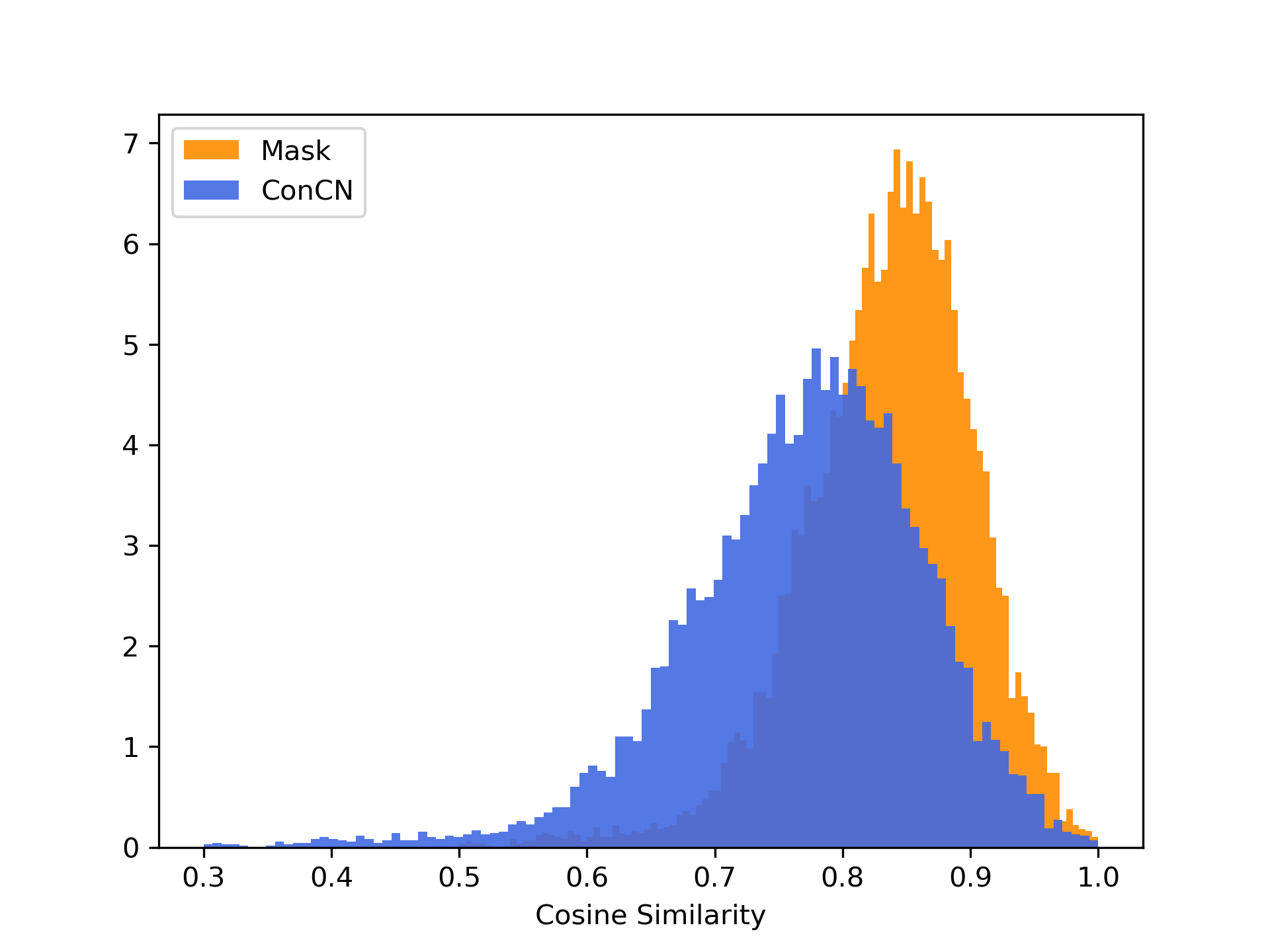}
\caption{Histogram of cosine similarities between the embeddings of two randomly sampled concepts, chosen from those appearing in the X-McRae, for the \emph{Mask} and \emph{ConCN}.\label{figAnisotropy1}}
\Description{This figure shows two histograms, one in blue and one in orange. The blue histogram shows the distribution of cosine similarities for the ConCN model; the orange histogram shows the distribution of cosine similarities for the Mask model.}
\end{figure}

\paragraph{Anisotropy}
As mentioned in the introduction, one of the reasons for the underperformance of the \emph{Mask} embeddings may be related to the high anisotropy of the BERT mention vectors. Figure \ref{figAnisotropy1} shows a histogram of the cosine similarities between randomly sampled concept embeddings, for the \emph{Mask} and \emph{ConCN} strategies. As can be seen, the cosine similarities are on average lower for \emph{ConCN}, which shows that this contrastive learning strategy has indeed led to a reduction in anisotropy.


\paragraph{Qualitative Analysis}
We now explore how the mention vectors are affected by the proposed fine-tuning strategy. Specifically, we consider pairs $(s_1,c_1), (s_2,c_2)$ where the mention vector for $(s_2,c_2)$ is among the top-100 nearest neighbours of the mention vector for $(s_1,c_1)$ when using \emph{ConCN}, while not being among the top-1000 nearest neighbours when using \emph{Mask} (for the full set of mention vectors $M$ across all words). Table \ref{tabSimilarForFinetuned} contains some examples of such sentence pairs.
The examples illustrate how fine-tuning allows the model to identify sentences that express similar properties, even when the sentences themselves are not similar, neither in syntactic structure nor in their overall meaning. In the first example, both sentences express that the target concept (which is masked) is some kind of building. Similarly, in the second example, the sentences express that the target concepts can be black. The third example illustrates a more abstract property, capturing the fact that country-specific versions of the target concept exist.  

Finally, Table \ref{tabNNs} shows the nearest neighbours of some selected target words, in terms of cosine similarity, for three different concept embeddings: \emph{Numberbatch}, \emph{MirrorBERT} and \emph{ConCN} (with filtering). For this analysis, we considered the vocabulary from the WordNet supersenses dataset. A first observation is that the neighbours for \emph{ConCN} are often taxonomically closer. For instance, for \emph{MirrorBERT} we see \emph{lemonwood} as a top neighbour of \emph{lemon}, which is topically related but not taxonomically close. Similarly, for both \emph{Numberbatch} and \emph{MirrorBERT} we see \emph{milk} as the second nearest neighbour of \emph{milkman}. As another difference, for \emph{ConCN} we can see neighbours which involve some abstraction. For instance, a \emph{button} has a similar role as an \emph{icon} in graphical user interfaces. Another notable example is \emph{cow} as a neighbour of \emph{milkman}, which are both related to the production/delivery of milk. However, this notion of abstraction sometimes also leads to sub-optimal neighbours. For instance, \emph{contradiction} is shown as one of the top neighbours of \emph{paradox} for both \emph{Numberbatch} and \emph{MirrorBERT} but does not appear as a neighbour for \emph{ConCN}.

\section{Conclusions}
We have proposed a method for learning concept embeddings, based on contextualised representations of masked mentions of concepts in a text corpus. Our focus was on improving the contextualised representations that can be obtained from a pre-trained BERT model, using a number of strategies based on contrastive learning. The aim of these strategies is to ensure that two contextualised word embeddings are similar if and only if the corresponding sentences express similar properties. To implement this idea, we need examples of sentences that are likely to express the same property. We have proposed two methods for obtaining such examples: an unsupervised method which relies on the neighbourhood structure of contextualised word vectors, and a distantly supervised method which relies on ConceptNet. In our experimental results, we found the latter method to perform best. Our proposed strategy was also found to outperform a range of baselines, both in word classification experiments and for the task of ontology completion.

\begin{acks}
This work was supported by ANR-22-CE23-0002 ERIANA and EPSRC grant EP/V025961/1. Na Li is supported by Shanghai Big Data Management System Engineering Research Center Open Funding. 
\end{acks}


\bibliographystyle{ACM-Reference-Format}
\balance
\bibliography{refs,anthology}

\end{document}